\documentclass[11pt]{article}

\usepackage[hyphens]{url}
\usepackage{hyperref}
\usepackage[hyphenbreaks]{breakurl}
\hypersetup{breaklinks=true}
\usepackage{emnlp2021}

\usepackage{times}
\usepackage{latexsym}

\usepackage[T1]{fontenc}
\usepackage[utf8]{inputenc}

\usepackage{microtype}
\usepackage{todonotes}
\usepackage[margin=10pt]{subfig}
\usepackage{float}
\usepackage{graphicx}
\usepackage{booktabs}
\usepackage{xcolor}

\title{How Emotionally Stable is ALBERT?\\ Testing Robustness with Stochastic Weight Averaging on a Sentiment Analysis Task}

 \author{Urja Khurana$^{\heartsuit}$ \and Eric Nalisnick$^{\diamondsuit}$ \and Antske Fokkens$^{\heartsuit\clubsuit}$ \\
$\heartsuit$ CLTL, Dept. of Language, Literature \& Communication, Vrije Universiteit Amsterdam\\$\diamondsuit$  Informatics Institute, University of Amsterdam\\
$\clubsuit$ Dept. of Mathematics and Computerscience, Eindhoven University of Technology\\
\texttt{\{u.khurana,antske.fokkens\}@vu.nl} \\ \texttt{e.t.nalisnick@uva.nl}}

\begin{document}
\maketitle
\begin{abstract}
Despite their success, modern language models are fragile. Even small changes in their training pipeline can lead to unexpected results. We study this phenomenon by examining the robustness of ALBERT \citep{lan2019albert} in combination with Stochastic Weight Averaging (SWA)---a cheap way of ensembling---on 
a sentiment analysis task (SST-2). In particular, we analyze SWA's stability via CheckList criteria \citep{ribeiro-etal-2020-beyond}, examining the agreement on errors made by models differing only in their random seed.
We hypothesize that SWA is more stable because it ensembles model snapshots taken along the gradient descent trajectory. We quantify stability by comparing the models' mistakes with Fleiss' Kappa \citep{fleiss1971measuring} and overlap ratio scores. We find that SWA reduces error rates in general; yet the models still suffer from their own distinct biases (according to CheckList). 
\end{abstract}

\section{Introduction}
Current language models perform well on data that resemble the distribution they are trained on, but even a slight variation in the model training setup can lead to results that diverge from what is originally reported \citep{fokkens-etal-2013-offspring, sellam2021multiberts}. Furthermore, when the model relies on spurious correlations for decision making, it then contains biases that are not represented by real world data. Ideally a model should be robust to data that has (slightly) different characteristics from the data it was trained on.  
 Accuracy and related metrics, despite their popularity, are usually not sufficient to identify these frailties. This is known as \textit{underspecification} \citep{d2020underspecification}: different predictors can achieve similar results on a specific task, but exhibit diverging performance on other tasks due to different induced biases. 

\textit{Stress tests} are an increasingly popular method for %which are evaluation tests that 
exposing biases of a model.
To test the linguistic capabilities and robustness of models, \citet{ribeiro-etal-2020-beyond} introduce \textit{CheckList}, an evaluation methodology that is comparable to the aforementioned \textit{stress tests} for robustness. CheckList can be used to investigate which linguistic phenomena are fully captured by a model and for which the model is thus expected to be robust across datasets.

Robustness and generalization can be improved by ensembling multiple models.
Training different models, however, is expensive. \textit{Stochastic Weight Averaging} (SWA) \citep{izmailov2018averaging} is a way of ensembling without the need to train different models. During training, the weights of the model at specific timepoints are averaged, avoiding the need to keep track of several models. The idea is that SWA explores different solutions close to a high performing minimum. 

In this paper, we study the effect of SWA on the robustness to both a standard sentiment analysis dataset and different CheckList capabilities. We investigate if models varying only in their random seeds still have different behavior on the same data when trained using SWA. Specifically, we train ALBERT-large \citep{lan2019albert} on SST-2 \citep{socher2013recursive}, a sentiment analysis dataset, with 10 random seeds.  We perform one run with SWA turned off (termed \textit{vanilla models}) and repeat the procedure with SWA turned on (termed \textit{SWA models}). We explore the robustness of the trained models using the CheckList methodology by looking at the stability of mistakes. We quantify this stability to measure the agreement in mistakes between the different models and compare the resulting values between the vanilla and SWA models. 

Our main hypothesis is that \textbf{using SWA leads to more stable models}. We therefore expect more overlap across random seeds in the results on the SST-2 evaluation data. We also expect SWA to lead to more overlap in mistakes for CheckList items that are captured by part of the vanilla models. We also anticipate (minor) improvements of general performance in both cases. For CheckList phenomena that are already largely captured or not at all, on the other hand, we do not expect to see major differences between vanilla models and SWA in terms of general performance or overlap.

\textbf{We make the following contributions}: 
\begin{itemize}
    \item We explore the effects of SWA on the stability and robustness of ALBERT-large that stem from underspecification. 
    \item We perform the, to our knowledge, first joint study of SWA and CheckList. 
    \item We provide an in-depth analysis of results by going beyond accuracy to look at overlap and agreement between random seeds and CheckList.
    \item We quantify agreement between different models by calculating overlap ratio and Fleiss' Kappa score on their mistakes.
\end{itemize}

We find that SWA improves on error rates in general, but results on increased stability are mixed: models with different random seeds still hold onto their own distinct induced biases on linguistic information captured by part of the models in our CheckList evaluation. There is minor improvement in stability on the Fleiss' Kappa score on the development set of SST-2, but results are not conclusive. Finally, we observe a large error rate for one of the random seeds on both SST-2 and CheckList, which also leads to a less strong result on increasing agreement between models.

\section{Related Work} 
\label{sec:related_work}
To the best of our knowledge, we are the first to combine SWA with CheckList and apply it to a BERT-based model to understand its effect on robustness with different random seeds. The work closest to ours has used variations of SWA for investigating the differences in interpretability on CNNs and LSTMs among different random seeds \citep{madhyastha-jain-2019-model}. A similar method to Stochastic Weight Averaging was employed by \newcite{xu2020improving} with a different objective: improving the fine-tuning process of BERT. They propose to average the BERT model at each time-step and two types of knowledge distillation to improve fine-tuning of the model. The averaging receives slightly better results and their variant of knowledge distillation works the best. However, it is unclear what the effect of this is on different random seeds. 

Instead of looking at a form of ensembling, \newcite{hua-etal-2021-noise} investigate the effect of injecting noise in BERT as a regularizer on the stability (sensitivity to input perturbation) of the models and show that fine-tuning performance improves. They point out that this improves generalizability as well, by looking at the difference in accuracy on the training and test set. However, training and test set might contain the same biases and hence might not reveal generalization issues \cite{elangovan-etal-2021-memorization}. 

\paragraph{Varying Performance} Most work until now has focused on behavioral changes of models on train and test data when changing an arbitrary choice of the pipeline, such as the random seed \citep{zhong2021larger, sellam2021multiberts}. Investigating the behavior of language models with different pre-training and fine-tuning random seeds on an instance-level, \citet{zhong2021larger} find that the fine-tuning random seed is influential for the variation in performance on an instance-level. This contrast in performance is also highlighted by \citet{sellam2021multiberts}; they release multiple BERT checkpoints with a different weight initialization and show diverging performance between similarly trained models. Such behavior has also been observed for out-of-distribution samples \citep{mccoy-etal-2020-berts, d2020underspecification, amir-etal-2021-impact}, where different induced biases are found when the random seed is modified and checkpoints behave differently on unseen data, even when evaluation performance is similar.  \citep{d2020underspecification, amir-etal-2021-impact}. \citet{watson2021agree} show that outputs from explainability methods also vary when changing hyperparameters, e.g.\ the random seed. 

\paragraph{Model Evaluation} Evaluating models on a development set might not expose certain biases or weaknesses a model has acquired due to the possibility of the same biases occurring in the training set. Hence, scalable diagnostic methodologies are useful to investigate a model's capabilities \citep{wu-etal-2019-errudite, ribeiro-etal-2020-beyond, wu2021polyjuice, goel2021robustness}. Even though these methodologies all focus on evaluation, the approach can vary between the methods. \citet{wu2021polyjuice} tackle evaluation from a counterfactual point of view. \citet{wu-etal-2019-errudite} not only examine counterfactuals but also grouping queries to ensure that error analysis is scaled to all instances. Likewise, \citet{goel2021robustness} exploit such subpopulation grouping, in addition to adversarial attacks, perturbations, and evaluation sets. It is possible to be unaware of certain subpopulations for which the model is weak, and therefore \citet{deon2021spotlight} introduce a method that looks for such weak groups. \citet{ribeiro-etal-2020-beyond} provide a methodology to analyze robustness toward basic capabilities and operationalize this with different test types (e.g.\ invariance to specific perturbations, basic capabilities). There are also more task specific efforts for evaluation, such as perturbations for robustness in task-oriented dialog \citep{liu-etal-2021-robustness} and evaluation of bias in a sentiment analysis setting \citep{asyrofi2021biasfinder}.

\section{Method} 
\label{sec:method}
\noindent To examine \textit{Stochastic Weight Averaging}'s effect on model stability due to \textit{underspecification}, we finetune a pretrained ALBERT-large version 2 on the SST-2 dataset. We train two types of models 10 times:\footnote{We provide all code at \url{https://github.com/cltl/robustness-albert}} 
\begin{itemize}
    \item \textit{Vanilla model}: Model finetuned with the hyperparameter values from \citet{lan2019albert}
    \item \textit{SWA model}: Model finetuned for the first few epochs with the hyperparameter values from \citet{lan2019albert} and then switching to a SWA training schedule
\end{itemize}

For all models, we keep the training protocol the same except for the random seed. 
 We train 10 models with a different random seed per model type. This gives us 20 different models: 10 \textit{vanilla models} and 10 \textit{SWA models}.\footnote{The experiments originally contained five random seeds, of which \textit{Random Seed 0} had exceptionally poor performance of $90.83$ accuracy on the development set. This was far from the reported validation accuracy of $94.9$ (\url{https://github.com/google-research/albert\#albert}). For the camera-ready version, we trained an additional five seeds, which confirmed that the anomalous one is indeed an outlier.} We then investigate the robustness of each model on CheckList tests and compare the performance of \textit{vanilla models} with \textit{SWA models}.

Due to \textit{underspecification}, the vanilla models are expected to have deviating performances on the tests across different random seeds, while the SWA models are expected to dampen this effect. We make a distinction between the following scenarios and what we expect: 
\begin{enumerate}
    \item \textbf{Linguistic information captured by all of the models}: We expect all of the models, regardless of the random seed, to be able to perform well on basic capabilities. Hence, we do not expect SWA to make much improvement, as there should not be a different behavior across random seeds. Stability will stay consistent here. 
    \item \textbf{Linguistic information captured by a part of the models}: This type of linguistic information is only captured by a part of the models due to their own induced biases. Hence, we expect that not all vanilla models behave similarly on such instances. With the introduction of SWA, more stability thus more overlap between mistakes is expected. 
    \item \textbf{Linguistic information captured by none of the models}: Some information cannot be captured by the model at all or it is unlikely that the model will be able to handle such information properly. In such cases, we do not expect SWA models to have an increase in performance, though that cannot be ruled out since it is possible that the weight space averaged by SWA is able to capture it. For the former, we do expect a large overlap of mistakes with SWA models since such information is not captured by any of the models.
\end{enumerate}

\subsection{Stochastic Weight Averaging}
\textit{Stochastic Weight Averaging} (SWA) is a cheap approach to create ensembles by averaging over different snapshots over the SGD trajectory, in contrast to the widely used approach of training different models \citep{izmailov2018averaging}. In essence, SWA ensembles in weight space instead of the usual model space. Due to the ensembling nature of correlated members from the same trajectory, we expect better generalization; a reduction in error rate and more stability in mistakes on unseen data.

We employ a strategy where the SWA models are trained in the same manner as the vanilla models for the first two epochs. This cut-off epoch is chosen empirically, by observing that the vanilla models start converging around 2-3 epochs. We make use of the Adam optimizer instead of the SGD optimizer since the former optimizer is used for the training of ALBERT. From the third epoch, the learning rate drops to a constant learning rate and at every end of the epoch, the model weights are averaged with the running average weights. With a high constant learning rate, the model is able to explore other solutions that are close to the local minimum that was found after two epochs and close to convergence. The respective constant learning rates of each random seed can be found in Table~\ref{tab:rs_lr}. The values for the learning rates are found empirically on the development set with the following candidate learning rates: \{6e-06, 7.5e-06\}.\footnote{We looked at the learning rates in examples from the original paper at \url{https://github.com/timgaripov/swa\#examples} where some SWA learning rates are half of the original learning rate and explored close candidate learning rates. From previous initial experiments learning rate $5e-06$ did not work and was thus left out in these sets of experiments.}

\begin{table}[h!]
    \centering
    \begin{tabular}{cc}
    \toprule
    & \textbf{SWA Learning Rate} \\
    \midrule
    Random Seed 0 & 6e-06 \\
    Random Seed 1 & 7.5e-06 \\
    Random Seed 2 & 6e-06 \\
    Random Seed 3 & 6e-06 \\
    Random Seed 4 & 7.5e-06 \\
    Random Seed 5 & 6e-06 \\
    Random Seed 6 & 6e-06 \\
    Random Seed 7 & 6e-06 \\
    Random Seed 8 & 6e-06 \\
    Random Seed 9 & 7.5e-06 \\
    \bottomrule
    \end{tabular}
    \caption{The constant SWA learning rates for each random seed.}
    \label{tab:rs_lr}
\end{table}

\subsection{SST-2 Dataset} 
We use the binary version of the Stanford Sentiment Treebank dataset\footnote{\url{https://nlp.stanford.edu/sentiment/index.html}} \citep{socher2013recursive}, which consists of human-annotated sentences from movie reviews originating from \url{rottentomatoes.com} for a sentiment classification task. This version of the dataset is also included in the GLUE task \citep{wang-etal-2018-glue}. We use this dataset since sentiment analysis is an interesting task to study underspecification as it is a more subjective task, making rigorous, multifaceted evaluation even more important. The training set consists of 67349 phrases, while the validation and test dataset consist of 872 and 1821 sentences respectively. We use the training and validation set for the training procedure, while the test set is used for the generation of specific CheckList items. 

\subsection{Checklist Evaluation} 
CheckList is a methodology to test basic and linguistic capabilities of a model, similar to behavioral testing in software engineering \citep{ribeiro-etal-2020-beyond}. They make a distinction between three types of tests:

\noindent \textbf{Minimum Functionality Test (MFT)}: Small examples to test for basic capabilities. We test if each instance has the specified label.\\ 
    \textbf{Invariance Test (INV)}: Tests that apply perturbations to the input and expect the prediction to stay consistent, \textit{regardless of the correctness of the prediction}. The original input together with its perturbations is seen as one test case.\\
\textbf{Directional Expectation Tests (DIR)}: Tests where the output is expected to change in a specific way, when the input is modified: the confidence is expected to change in a specific direction. Similar to INV tests, the original input with modifications is seen as a test case.

In this paper, we consider different MFTs, INVs, and DIRs tests for sentiment analysis. We check for basic capabilities and robustness. Each trained model is evaluated on our CheckList set up and their performances are compared. We expect that vanilla models make more mistakes than SWA models and qualitatively make less overlapping mistakes due to each model having their own different induced biases. On the other hand, SWA models are expected to have more overlapping mistakes, due to its ensembling and explorative nature in the weight space. 

We created 18 CheckList capability tests by adapting tests from the CheckList GitHub repository\footnote{\url{https://github.com/marcotcr/checklist/blob/master/notebooks/Sentiment.ipynb}} to the use-case in this paper. For reasons of space, we refer to individual capability tests with transparent names followed by the test size, only using short explanations when the name by itself is not sufficiently clear. For tests that perturb the input and are not created from scratch, we use the test set from SST-2. Each original input can be augmented more than once, depending on the capability. These tests are followed by two numbers when introduced: the number of original items and total items with perturbations included. A full overview of the CheckList capabilities and their sizes can be found in Table~\ref{tab:checklist_tests} in Appendix~\ref{appendix:checklist_tests}.

\section{Results} 
\label{sec:results}

This section presents the outcome of our experiments. We first provide results on the original dataset and then the results on CheckList items. Lastly we examine how stable vanilla and SWA models are by looking at the label agreement between models trained from different seeds.

\subsection{Stochastic Weight Averaging}

As mentioned in the previous section, we originally ran our experiments on five random seeds and added five additional seeds after observing that one seed performed lower than all others. When we compare the accuracy of the vanilla models with the SWA models on the validation set of SST-2 in Table~\ref{tab:model_val_accuracy}, it is evident that most of the SWA models perform slightly better than the vanilla models. The only exceptions are \textit{Random Seed 0, 7,} and \textit{8}. Upon running our experiments on five additional seeds, \textit{Random Seed 0} remains the only seed that has an accuracy around $0.90$, confirming that it is an outlier. The SWA versions of the other two random seeds might not outperform their vanilla counterparts but achieve a close accuracy.

Due to the outlying behavior of \textit{Random Seed 0}, we leave its results out of the rest of the analysis, to avoid noise from this model influencing the analysis. We
present the complete results with \textit{Random Seed 0} included in Appendix~\ref{appendix:results_with_rs0}.

\begin{table}[h!]
    \centering
    \begin{tabular}{ccc}
        \toprule
        & \textbf{Vanilla} & \textbf{SWA} \\ 
        \midrule
        Random Seed 0 & \textbf{0.9083} & 0.8991 \\
        Random Seed 1 & 0.9507 & \textbf{0.9541} \\
        Random Seed 2 & 0.9450 & \textbf{0.9495} \\
        Random Seed 3 & 0.9507 & \textbf{0.9541} \\
        Random Seed 4 & 0.9450 & \textbf{0.9461} \\
        \midrule
        Random Seed 5 & 0.9495 & \textbf{0.9507} \\
        Random Seed 6 & 0.9450 & \textbf{0.9472} \\ 
        Random Seed 7 & \textbf{0.9438} & 0.9392 \\
        Random Seed 8 & \textbf{0.9461} & 0.9450 \\ 
        Random Seed 9 & 0.9415 & \textbf{0.9461} \\
        \bottomrule
    \end{tabular}
    \caption{Accuracy on the validation set of SST-2 for the vanilla and SWA models of the different random seeds.}
    \label{tab:model_val_accuracy}
\end{table}

\subsection{CheckList Evaluation}
\subsubsection{Vanilla Model Results}

\begin{figure*}[h!]
    \centering
    \subfloat[Error rates of each vanilla random seed for each CheckList capability. ]{\includegraphics[width=\textwidth]{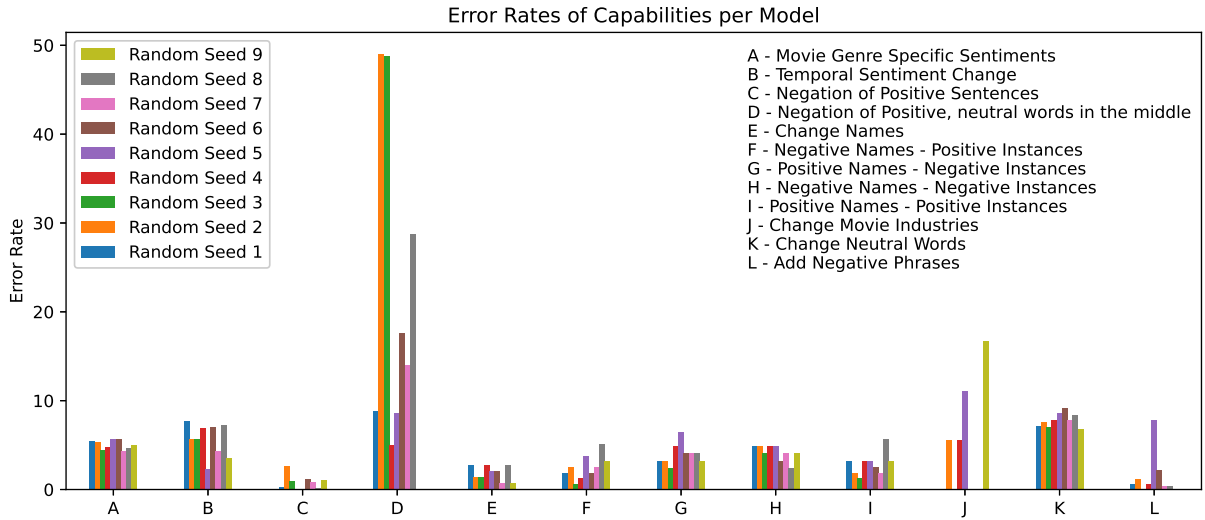}\label{fig:checklist_error_vanilla}}\\
    \subfloat[Error rates of each SWA random seed for each CheckList capability.]{\includegraphics[width=\textwidth]{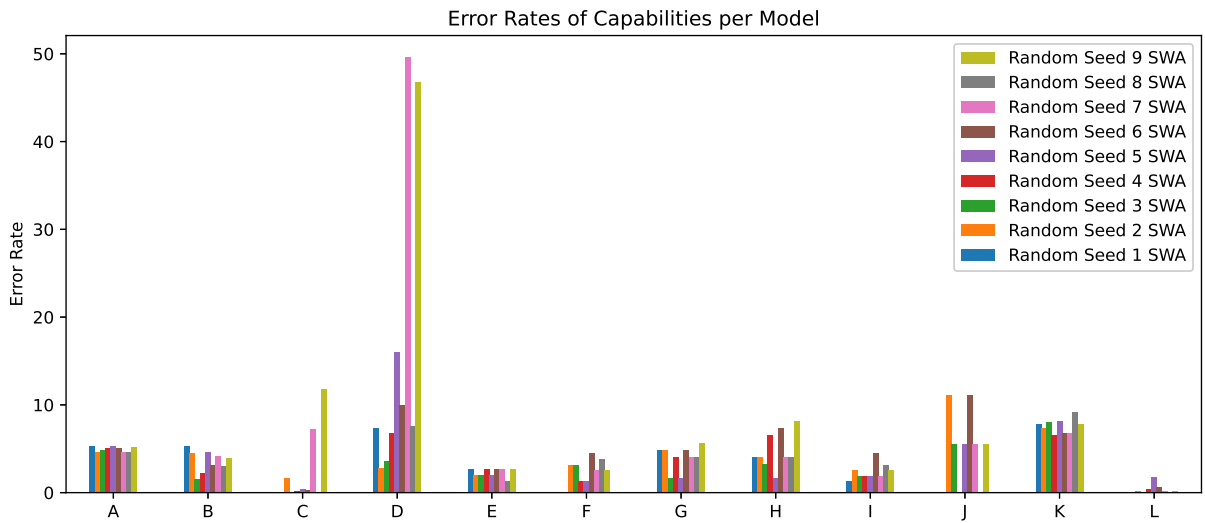}\label{fig:checklist_error_swa}}
    \caption{Comparison of error rates per capability of vanilla and SWA models.}
    \label{fig:checklist_error_rates}
\end{figure*}

\paragraph{Error Rates} We show the failure rate for each capability per vanilla model in Figure \ref{fig:checklist_error_vanilla}. For the \textit{Movie Sentiments} (n=58), \textit{Single Positive Words} (n=22), \textit{Single Negative Words} (n=14), and \textit{Sentiment-laden Words in Context} (n=1350) capabilities there are no mistakes made by any of the vanilla models. On \textit{Add Positive Phrases} (n=500, m=5500), only \textit{Random Seed 8} makes mistakes with a very small error rate. Similarly, on \textit{Movie Industries Sentiments} (n=1200) only \textit{Random Seed 8} and \textit{Random Seed 2} make mistakes, again with very small error rates that would not be visible on the plot. Hence for clarity, these capabilities are left out of the plot.

There is not much variation in the error rate for most of the capabilities. The most variation in performance among the random seeds can be observed for the capability that tests negations of positive sentences, with a neutral sentiment in the middle of the sentence: \textit{Negation of Positive, neutral words in the middle} (D) (n=500). Interestingly, it is evident that particular random seeds can deal with negation better than others: \textit{Random Seed 1, 4,} and \textit{5}. These random seeds have the lowest error rate for both \textit{Negation of Positive Sentences} (C) (n=1350) and \textit{Negation of Positive, neutral words in the middle}.

\paragraph{Overlap Ratios} A similar error rate, however, does not mean that the errors occur for the same instances. Hence, we analyze the overlap of errors of the vanilla models per capability. We calculate an overlap ratio by dividing the intersection of the failures of two random seeds by the union of those same failures. In contrast to the error rates, the overlap ratios are on an instance-level instead of case-level. There is no overlap of errors between the models for the capability \textit{Add Positive Phrases}. The capability with the highest overlap ratio is \textit{Movie Genre Specific Sentiments} (A) (n=736), which checks for sentiments that are fitting or not for specific genres: e.g.\ a scared feeling after watching a horror movie. This indicates that most of the models make similar mistakes for this capability. When looking at the mistakes, all the models misclassify sentences about horror movies being terrifying, scary, frightening or calming, a comedy movie being serious and a drama movie being funny instead of serious. In general, most of the vanilla models have a low overlap ratio, with the only exceptions being \textit{Negation of Positive, neutral words in the middle} (D) and \textit{Temporal Sentiment Change} (B) (n=2152). The latter capability contains sentences where the sentiment changes over time. These two contain certain random seeds that achieve a higher overlap ratio, as we can see in the spread of the box for these capabilities.

\subsubsection{SWA Model Results}

\begin{figure*}[h!]
    \centering
    \subfloat[Comparison of variation in error rates between vanilla (red boxes) and SWA models (blue boxes), showcased per CheckList capability. Outliers are indicated with a circle.]{\includegraphics[width=\textwidth]{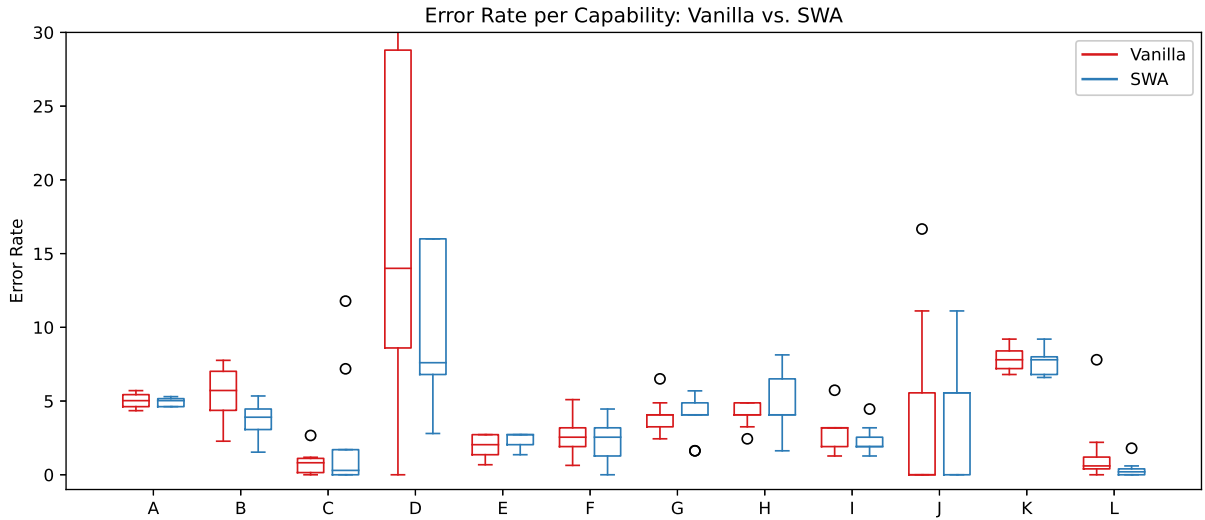}\label{fig:error_rate_boxplot}} \\
    \subfloat[Comparison of variation in overlap ratios between vanilla (red boxes) and SWA models (blue boxes), showcased per CheckList capability. Outliers are indicated with a circle.]{\includegraphics[width=\textwidth]{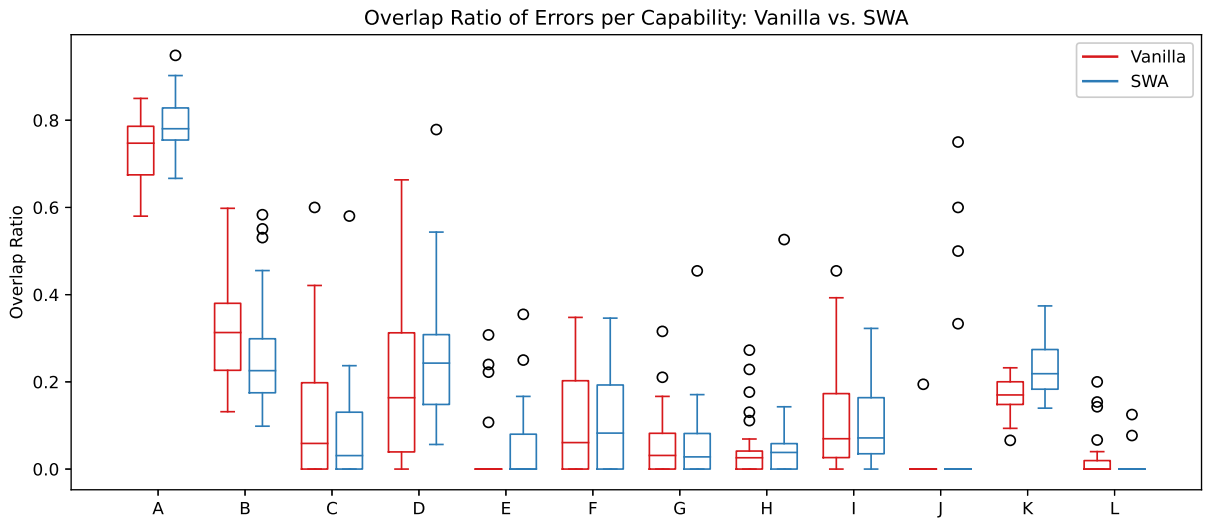}\label{fig:overlap_ratio_boxplot}}
    \caption{Comparing the variation of error rates and overlap ratios per capability for vanilla and SWA models. Legend for the x-axis can be found in Figure \ref{fig:checklist_error_rates}}
\end{figure*}
\paragraph{Error Rates} \noindent Error rates for the SWA models per capability can be found in Figure~\ref{fig:checklist_error_swa}. In general, we can observe a (slight) reduction in error rate with SWA models compared to vanilla models. On \textit{Add Positive Phrases}, only \textit{Random Seed 5} and \textit{Random Seed 6} have a slight increase in error rate. The latter is also the only one to make a mistake on \textit{Movie Industries Sentiments}\footnote{These error rates are too low (0.2\% - 0.35\%) to be visible in the plot, hence left out.}. The largest drop can be seen for \textit{Negation of Positive, neutral words in the middle} (D), where the diverging performance seen for the vanilla models has been reduced for most random seeds, except for \textit{Random Seed 7} and \textit{Random Seed 9}, whose error rates increase significantly. Similar behavior can be observed for \textit{Negation of Positive Sentences} (C), where only the SWA versions of \textit{Random Seed 7} and \textit{9} have an increase in error rate. This suggests that the SWA solution for these two random seeds is worse in handling negation than their corresponding vanilla versions. For other capabilities, the error rate mostly reduces slightly or stays the same. The only exceptions are \textit{Positive Names - Negative Instances} (G) (n=123, m=1353) and \textit{Negative Names - Negative Instances} (H) (n=123, m=1353), where Negative Names are names that tend to occur in negative reviews in the training data, similarly for positive names, and we insert these names in negative instances of the test set. More details are provided in Appendix~\ref{appendix:checklist_tests}. 

\paragraph{Overlap Ratios} The overlap ratio for most capabilities remains low. Notably, the spread of overlap ratio for \textit{Movie Genre Specific Sentiments} (A) increases from the vanilla models. All of the models still struggle with understanding that horror movies being terrifying, scary or frightening is positive, and calming is negative. This is in line with the expectations of SWA not improving (much) on capabilities that are not captured by any of the models. We find increase in overlap for \textit{Change Names} (E) (n=147, m=1617), \textit{Negative Names - Negative Instances} (H), and \textit{Change Neutral Words} (K) (n=500, m=3846), in accordance with our expectation of SWA bringing more stability. There is a different trend, against expectations, for \textit{Add Negative Phrases} (L), \textit{Negation of Positive Sentences} (C), and \textit{Temporal Sentiment Change} (B), where the large variation of overlap of vanilla models is reduced significantly. For the rest of the capabilities, the overlap ratio appears to stay somewhat the same. 

Overall, there are three different outcomes when comparing stability with SWA to vanilla models: (1) Good performance of vanilla models stays consistent for the SWA models. (2) Large variations in error rates with vanilla models are reduced with SWA, but the overlap of mistakes does not increase and might decrease for some cases. (3) Overlap ratio with SWA does not necessarily increase, when error rates of the vanilla models are somewhat similar and remain the same for the SWA models. As such, we do not find evidence to confirm our hypothesis based on overlap between the outcomes on CheckList items.

\subsection{Fleiss' Kappa}
To further investigate the stability of SWA models, we measure the inter-model agreement on the misclassifications with the use of Fleiss' Kappa \citep{fleiss1971measuring}. This measure is used for inter-annotator agreement which can be related to the nine random seeds. In our case the annotators and the predictions being their annotations, used for both the vanilla and SWA models. Negative values or values close to zero are considered to indicate a rather low agreement, while the higher the value, the more agreement there is.

\begin{table}[h!]
    \centering
    \scalebox{0.9}{
    \begin{tabular}{p{2cm}ccc}
        \toprule
         & \textbf{Vanilla} &  \textbf{SWA }& \textbf{Difference} \\
         \midrule
         \textit{With Random Seed 0} & 0.205964 & 0.247299 & 0.041335 \\[0.2cm]
         \textit{Without Random Seed 0} & 0.226725 & 0.360317 & 0.133592 \\
         \midrule
         \textit{With Random Seed 0} & 0.3984 & 0.4381 & 0.03967 \\[0.2cm]
         \textit{Without Random Seed 0} & 0.3881 & 0.4106 & 0.0225 \\
         \bottomrule
    \end{tabular}}
    \caption{Fleiss' Kappa values of the vanilla and SWA models on the agreement on the misclassifications on the development set. The upper block is with the first five random seeds and the lower is with all 10.}
    \label{tab:fleiss_kappa_dev_mistakes}
\end{table}

The results on the development set in Table~\ref{tab:fleiss_kappa_dev_mistakes} illustrate a significant increase in agreement for SWA models, when considering the initial four random seeds, without outlier \textit{Random Seed 0}. While the agreement is still on the lower side, hinting at the presence of induced biases, the increase indicates more agreement on errors between the models and lesser distinct mistakes. We hence look at the Fleiss' Kappa values with the additional five random seeds incorporated. The Fleiss' Kappa agreement increases significantly in general for both the vanilla and SWA models.
We now only observe a small increase in agreement when applying SWA compared to the vanilla models.

\begin{table}[h!]
    \centering
    \hspace*{-0.5cm}
    \scalebox{0.75}{
    \begin{tabular}{p{3.5cm}rrr}
        \toprule
        {} &   \textbf{Vanilla} &       \textbf{SWA} &  \textbf{Difference} \\
        \midrule
        Negation of Positive Sentences                    &  0.029640 &  0.020448 &   -0.009192 \\
        Negation of Positive, neutral words in the middle &  0.107637 &  0.142219 &    0.034582 \\
        Movie Genre Specific Sentiments                   &  0.581853 &  0.660138 &    0.078285 \\
        Temporal Sentiment Change                         &  0.248653 &  0.290926 &    0.042273 \\
        % Movie Industries Sentiments                       & -0.048387 & -0.036866 &    0.011521 \\
        \midrule
        Change Names                                      & -0.091694 & -0.084096 &    0.007598 \\
        Negative Names - Positive Instances               &  0.006975 &  0.006021 &   -0.000954 \\
        Positive Names - Negative Instances               & -0.069162 & -0.076226 &   -0.007064 \\
        Negative Names - Negative Instances               & -0.082486 & -0.069141 &    0.013346 \\
        Positive Names - Positive Instances               &  0.012704 &  0.035196 &    0.022492 \\
        Change Movie Industries                           & -0.072503 & -0.052239 &    0.020264 \\
        Change Neutral Words                              &  0.087306 &  0.135759 &    0.048453 \\
        \midrule
        Add Negative Phrases                              & -0.031328 & -0.062053 &   -0.030724 \\
        \bottomrule
    \end{tabular}}
    \caption{Fleiss' Kappa values of the vanilla and SWA models on the agreement on CheckList mistakes per capability. The first part of the table shows the MFT capabilities, the second part are the INV capabilities, and the third part are the DIR capabilities.}
    \label{tab:fleiss_kappa_checklist_mistakes}
\end{table}

We calculate the Kappa measure on the predictions of all the random seeds on the CheckList items as well. For the tests that measure basic capabilities (MFTs), we look at the agreement on predictions of errors. With tests that perturb an input (INVs), the instances that flip the output prediction are considered as a failure, so we check for model agreement on flipping for an instance. Similarly, for capabilities that test a directional change in confidence (DIRs), instances that go against the expected direction are considered failures and we compare model agreement on if they change in the same direction. 

The Kappa values for the CheckList mistakes in Table~\ref{tab:fleiss_kappa_checklist_mistakes} stay mostly unchanged with slight increases or decreases in agreement. This is in accordance with the results observed for the development set mistakes: it appears that SWA does not provide the stability across random seeds and still suffers from its own induced biases. Generally, the agreement is on the lower side. The Kappa values for \textit{Movie Industries Sentiments} and \textit{Add Positive Phrases} were $0.0$ for both vanilla and SWA models and hence left out of the table. For \textit{Movie Genre Specific Sentiments} we see a large agreement and the biggest increase in agreement with SWA. This corresponds to the high overlap ratio for the same capability.

While SWA globally cuts down on error rate, it appears that this does not necessarily translate to improvement in stability: there is still disagreement in the labels assigned by individual models. Even with SWA, the models appear to make different errors on CheckList as confirmed by the low Kappa values and overlap ratio. For some capabilities the spread of the overlap ratio is on the higher side, indicating that \textit{some} random seed models are closer to each other in terms of decision making, but this does not hold for \textit{all}.

\section{Discussion} 
\label{sec:discussion}
This research illustrates the potential impact of random seeds. First, our original sample of 5 seeds contained an outlier that performed far worse than the other seeds (as well as the original study). Second, while initial results on the SST-2 development set were promising when looking at the 4 random seeds that showed normal behavior, these results did not hold when adding 5 additional random seeds. This highlights the necessity for proper analysis and the fragility of deep language models. Possibly, the initial random seeds were closer to each other in the weight space and hence SWA appeared to increase the agreement significantly. The additional random seeds could lie farther away, thus subsiding the increased agreement. In the future, more comprehensive research on the proximity and behavior of different random seeds could therefore be useful.

Even though CheckList provides an easy way to investigate the capabilities of a model, automatizing some tests can be hard. There can be situations in which labels indicated for a specific capability might not hold for a certain test case. For instance, negating a negative sentence might not always lead to a positive sentence, it can also be neutral. Similarly, we applied negations on some instances from the test set but the label is not required to flip, depending on the placement of the negation. Therefore, we leave out the results in our conclusions as the labels did not always make sense upon investigation. In some instances, it is also unclear what the resulting label should be. We have added the results for these specific capabilities in Appendix~\ref{appendix:excluded_capabilities} for completeness. For further experiments, we would like to manually generate some CheckList capabilities to ensure validity of the labels. This will also enable us to focus on the creation of more subjective tests, cases that are less black-and-white than the tests conducted in this research. We can then gain more insights into the fragility of models when it comes to border cases.

\section{Conclusion} 
\label{sec:conclusion}
We combine SWA with the CheckList methodology to explore the effects of SWA on the robustness of a BERT-based model (ALBERT-large) on different random seeds and apply it to a sentiment analysis task. To understand how SWA affects the stability amongst different random seeds, we analyze in-depth the results and mistakes made on the development set and CheckList test items and provide error rates, overlap ratios, and Fleiss' Kappa agreement values. While SWA is able to reduce the error rate in general amongst most of the random seeds, on the CheckList tests, there are still some capabilities that models make their own distinct mistakes on with SWA incorporated. The stability on the development set also improves only slightly. In the future, we would like to create more hand-crafted CheckList capabilities for further rigorous study. Furthermore, it could be useful to thoroughly investigate the impact of adjacency of random seeds on their error agreement.

\section*{Acknowledgements}
This research was (partially) funded by the Hybrid Intelligence Center, a 10-year programme funded by the Dutch Ministry of Education, Culture and Science through the Netherlands Organisation for Scientific Research.

\bibliography{anthology,custom}
\bibliographystyle{acl_natbib}

% \newpage
\appendix
\section{Technical Details}
\label{appendix:training_details}
\begin{figure}[H]
    \hspace*{-0.5cm}
    \centering
    \includegraphics[width=0.5\textwidth]{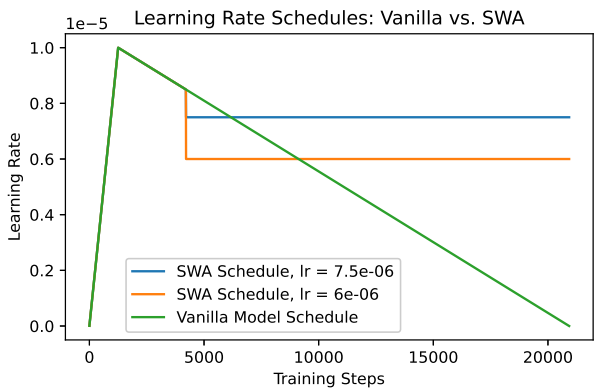}
    \caption{All the different learning rate schedules start identically with 1256 warmup steps to 1e-5. For the vanilla models (the green line) the learning rate anneals linearly to 0 till the 20935th step. This is in accordance with the hyperparameters reported in \citet{lan2019albert}. For the SWA models, after the second epoch, the learning rate drops to one of the specified learning rates (blue or orange lines) and stays constant.}
    \label{fig:lr_schedule}
\end{figure}

For model training, we make use of the HuggingFace \citep{wolf2019huggingface} pipeline and train the models on a single GeForce RTX 2080 Ti. We use the same hyperparameter settings as reported by \citet{lan2019albert}. The visualization of the learning rate schedules can be seen in Figure~\ref{fig:lr_schedule}.

As the HuggingFace pipeline does not provide the labels for the test set of SST-2, we match the phrases of the test set in HuggingFace with the phrases in the SST-2 dataset from the \textit{dictionary.txt} file, downloaded from GLUE,\footnote{\url{https://gluebenchmark.com/tasks}} to get their phrase IDs. Then we use those IDs to extract the labels from \textit{sentiment\_labels.txt}. Every label above $0.6$ is mapped to \textit{positive} and equal to or lower than $0.4$ is mapped to \textit{negative}, as mentioned in the instructions of the \textit{README.md} file. Some sentences are matched manually as they differ only in British vs. American English spelling.

\section{Results of Excluded Capabilities}
\label{appendix:excluded_capabilities}
For completeness, we also show the results for capabilities excluded from our analysis. For \textit{Add Negations} and \textit{Negation of Negative Sentences} we generated automatic test cases but the labels were not always correct upon investigation. Hence, we left these two capabilities out of the analysis. 

\begin{figure}[h!]
    \centering
    \subfloat[Error rates of each vanilla random seed for the excluded capabilities.]{\includegraphics[width=0.5\textwidth]{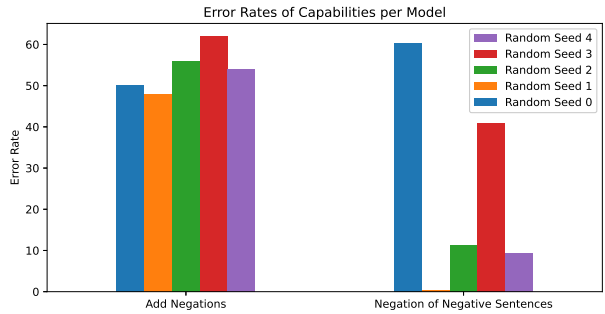}\label{fig:negations_error_vanilla}} \\
    \subfloat[Error rates of each SWA random seed for the excluded capabilities.]{\includegraphics[width=0.5\textwidth]{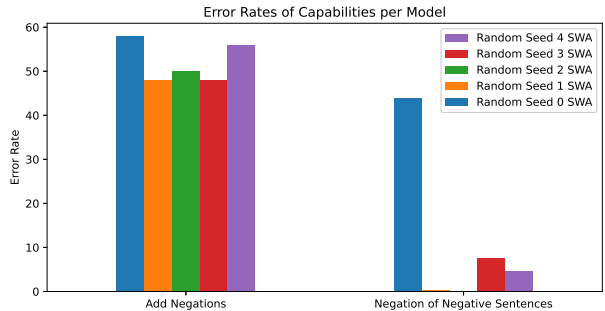}\label{fig:negations_error_swa}}
    \caption{Comparison of error rates of vanilla and SWA models for the excluded capabilities.}
\end{figure}

\begin{figure}[h!]
    \centering
    \subfloat[Comparison of variation in error rates between vanilla (red boxes) and SWA models (blue boxes), for the excluded capabilities.]{\includegraphics[width=0.5\textwidth]{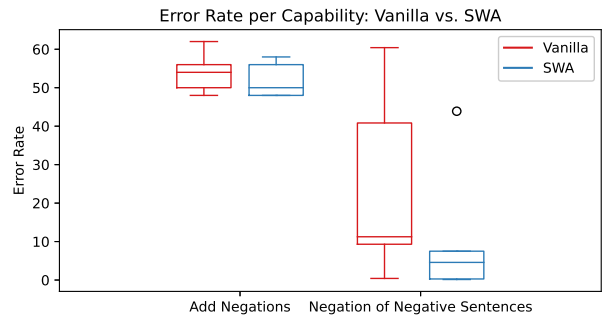}\label{fig:negations_error_boxplot}} \\
    \subfloat[Comparison of variation in overlap ratios between vanilla (red boxes) and SWA models (blue boxes), for the excluded capabilities.]{\includegraphics[width=0.5\textwidth]{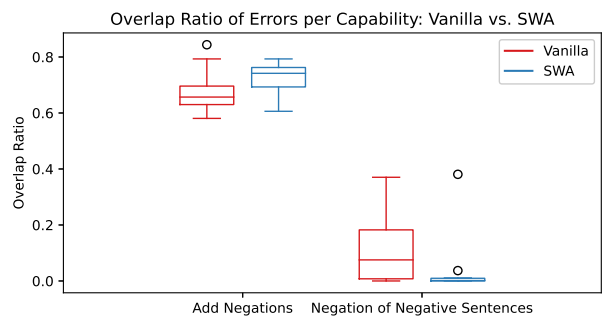}\label{fig:negations_overlap_ratio}}
    \caption{Comparing the variation of error rates and overlap ratios for the excluded capabilities. }
\end{figure}

In Table \ref{tab:fleiss_kappa_negatives} we show the Fleiss' Kappa values, the error rates per capability for the vanilla and SWA models can be found in Figure~\ref{fig:negations_error_vanilla} and Figure~\ref{fig:negations_error_swa}, respectively. The variation in error rates and overlap ratios between vanilla and SWA models can be found in the Figures~\ref{fig:negations_error_boxplot} and \ref{fig:negations_overlap_ratio} respectively. All the results are with the five initial random seeds, \textit{Random Seed 0} included. 

\vspace*{1cm}
\begin{table}[h!]
    \centering
    % \hspace*{-0.5cm}
    \scalebox{0.8}{
    \begin{tabular}{p{3cm}ccc}
        \toprule
        {} &   \textbf{Vanilla} &       \textbf{SWA} &  \textbf{Difference} \\
        \midrule
        Add Negations                  &  0.475553 &  0.470416 &   -0.005136 \\
        Negation of Negative Sentences & -0.025882 & -0.066728 &   -0.040845 \\
        \bottomrule
    \end{tabular}}
    \caption{Fleiss' Kappa values of the excluded capabilities. }
    \label{tab:fleiss_kappa_negatives}
\end{table}

% \newpage
\section{CheckList Results with Random Seed 0}
\label{appendix:results_with_rs0}
\begin{figure*}[h!]
    % \hspace*{-2.4cm}
    
    \centering
    \subfloat[Error rates of each vanilla random seed for each CheckList capability. ]{\includegraphics[width=0.9\textwidth]{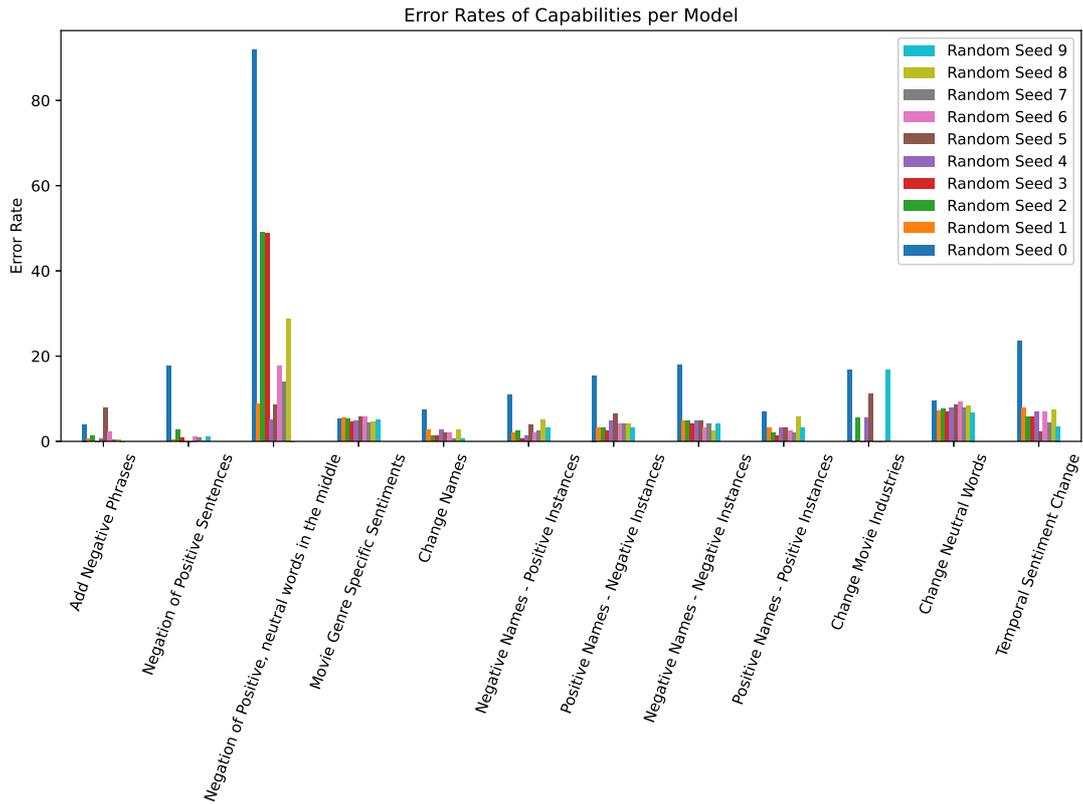}\label{fig:checklist_error_vanilla_with0}}\\
    \subfloat[Error rates of each SWA random seed for each CheckList capability.]{\includegraphics[width=0.9\textwidth]{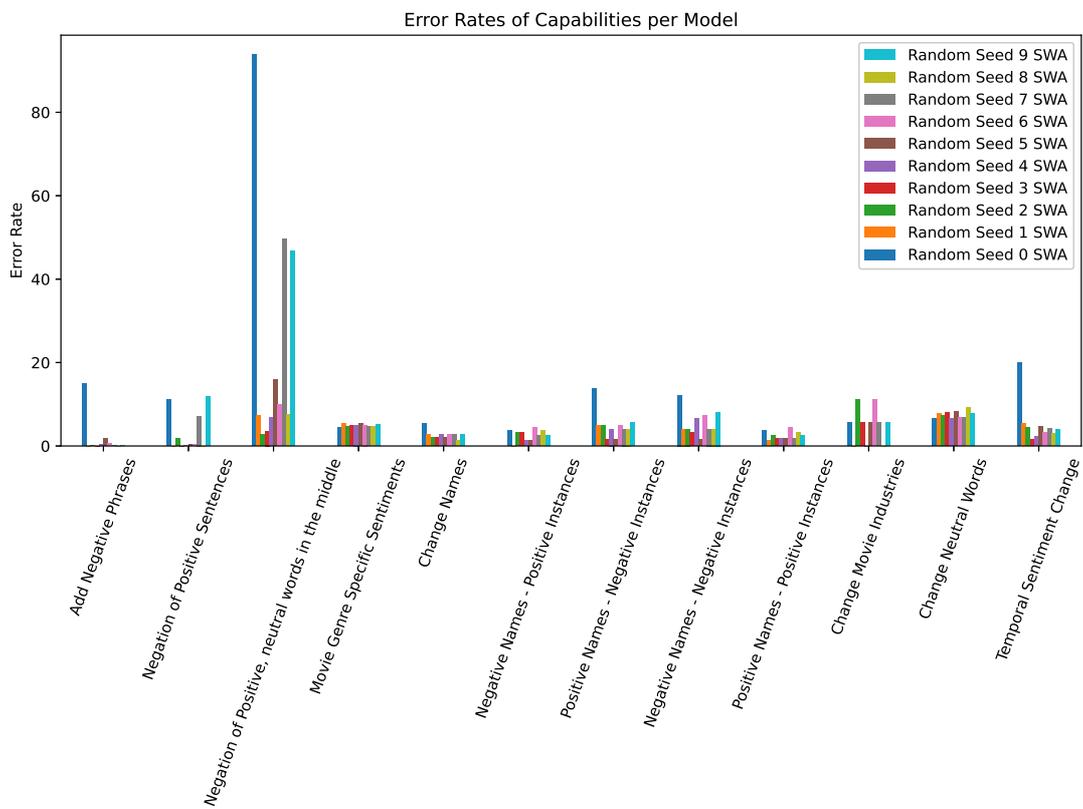}\label{fig:checklist_error_swa_with0}}
    \caption{Comparison of error rates per capability of vanilla and SWA models with all the 10 random seeds.}
\end{figure*}

\begin{figure*}[h!]
    \centering
        
    \subfloat[Comparison of variation in error rates between vanilla (red boxes) and SWA models (blue boxes), showcased per CheckList capability. Outliers are indicated with a circle.]{\includegraphics[width=0.9\textwidth]{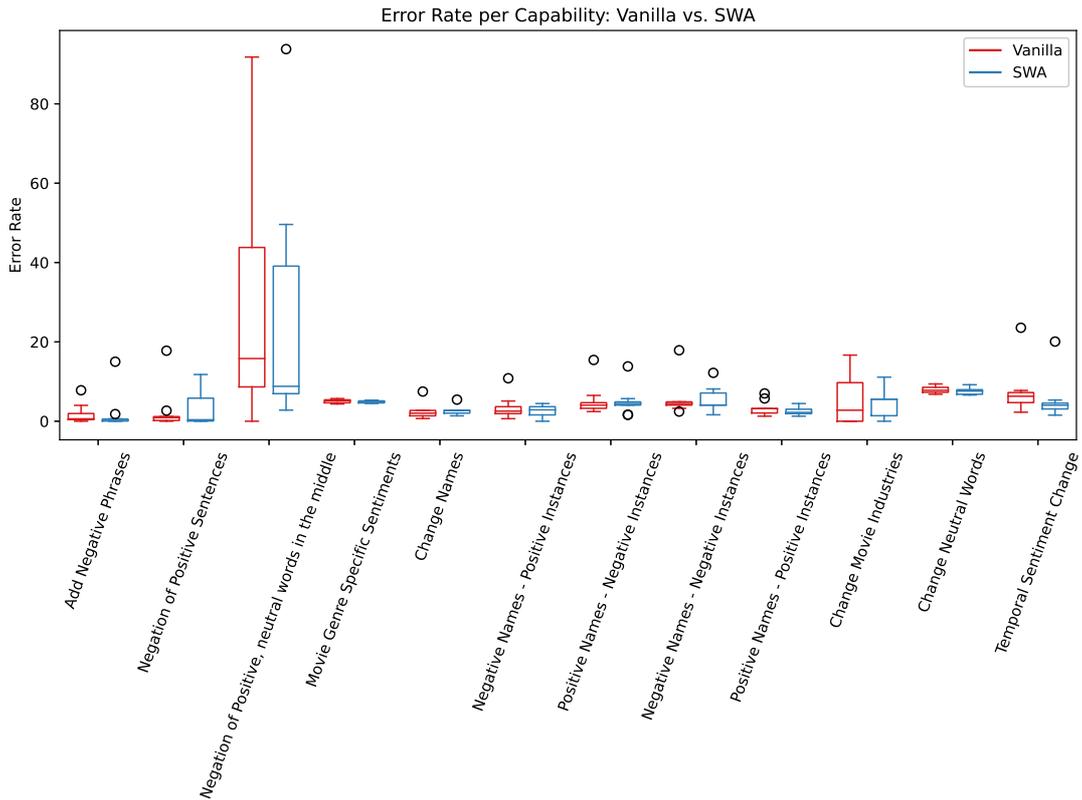}\label{fig:error_rate_boxplot_w0}} \\
    \subfloat[Comparison of variation in overlap ratios between vanilla (red boxes) and SWA models (blue boxes), showcased per CheckList capability. Outliers are indicated with a circle.]{\includegraphics[width=0.9\textwidth]{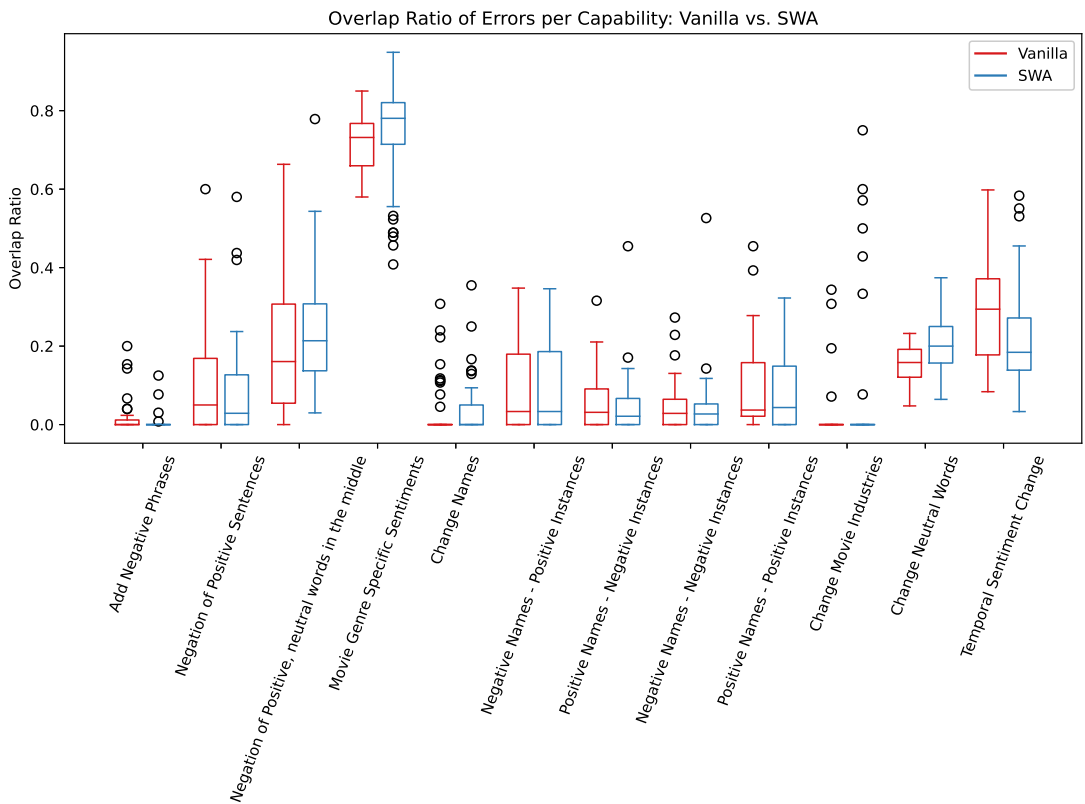}\label{fig:overlap_ratio_boxplot_w0}}
    \caption{Comparing the variation of error rates and overlap ratios per capability for vanilla and SWA models, including results from \textit{Random Seed 0}. }
\end{figure*}

We present our results on CheckList with \textit{Random Seed 0} as well for transparency. We again present the Fleiss' Kappa values for the CheckList capabilities in Table~\ref{tab:fleiss_kappa_checklist_mistakes_withrs0}. The error rates of each capability per vanilla and SWA models can be found in Figures~\ref{fig:checklist_error_vanilla_with0} and \ref{fig:checklist_error_swa_with0}. We also plot the variation in error rates (Figure~\ref{fig:error_rate_boxplot_w0}) and overlap ratios (Figure~\ref{fig:overlap_ratio_boxplot_w0}). 

\begin{table*}[hb!]
    \centering
    % \scalebox{0.8}{
    \begin{tabular}{cccc}
        \toprule
        {} &   \textbf{Vanilla} &       \textbf{SWA} &  \textbf{Difference} \\
        \midrule
        Negation of Positive Sentences                    &  0.048956 &  0.031777 &   -0.017179 \\
        Negation of Positive, neutral words in the middle &  0.132425 &  0.150507 &    0.018082 \\
        Movie Genre Specific Sentiments                   &  0.583256 &  0.590209 &    0.006952 \\
        Sentiment-laden Words in Context                  &  0.000000 &  0.000000 &    0.000000 \\
        Temporal Sentiment Change                         &  0.335020 &  0.428743 &    0.093723 \\
        Movie Industries Sentiments                       & -0.048387 & -0.036866 &    0.011521 \\
        \midrule
        Change Names                                      & -0.046527 & -0.074413 &   -0.027887 \\
        Negative Names - Positive Instances               & -0.007656 & -0.007270 &    0.000386 \\
        Positive Names - Negative Instances               & -0.044586 & -0.067630 &   -0.023044 \\
        Negative Names - Negative Instances               & -0.046272 & -0.061678 &   -0.015406 \\
        Positive Names - Positive Instances               &  0.001146 &  0.017189 &    0.016043 \\
        Change Movie Industries                           & -0.041426 & -0.042046 &   -0.000620 \\
        Change Neutral Words                              &  0.072667 &  0.124769 &    0.052101 \\
        \midrule    
        Add Positive Phrases                              & -0.058824 & -0.043796 &    0.015028 \\
        Add Negative Phrases                              & -0.051318 & -0.021295 &    0.030023 \\
        \bottomrule
    \end{tabular}
    % }
    \caption{Fleiss' Kappa values of the vanilla and SWA models on the misclassifications on the CheckList tests separately, all the 10 random seeds, including outlier \textit{Random Seed 0}. The first part of the table shows the MFT capabilities, the second part are the INV capabilities, and the third part are the DIR capabilities.}
    \label{tab:fleiss_kappa_checklist_mistakes_withrs0}
\end{table*}

\newpage
\section{CheckList Capabilities}
\label{appendix:checklist_tests}
In Table \ref{tab:checklist_tests} we describe each CheckList capability that we test for. For perturbing capabilities such as \textit{Negative names, Positive instances} and its other variants, we extract names from the SST-2 training set with Spacy \citep{spacy}. Due to false positives, we manually remove names that do not refer to a person, such as movie names and historical figures. Per name, we calculate the mean of labels of the instances it occurs in. This way, we can select positive and negative names to perturb test set instances with.

As reviews were predominantly about Hollywood, we also perturbed instances talking specifically about it. We compile a list of around 10 other movie industries,\footnote{\sloppy{\url{https://en.wikipedia.org/wiki/List_of_Hollywood-inspired_nicknames}}} based on how many movies are produced\footnote{\sloppy{\url{https://en.wikipedia.org/wiki/Film_industry\#Statistics}}} and revenue.

\begin{table*}[h!]
    \centering
    \scalebox{0.75}{
    \begin{tabular}{p{4cm}ccp{7cm}p{3cm}}
        \toprule
        \textbf{Capability} & \textbf{Test Type} & \textbf{\#Examples} & \textbf{Description} & \textbf{Original CheckList Name} \\
        \midrule
        Single Positive Words & MFT & 22 & Positive words that should be predicted \textit{positive} (e.g. beautiful, brilliant, enjoyed).\\
        Single Negative Words & MFT & 14 & Negative words that should be predicted \textit{negative} (e.g. hate, disliked, dreaded). \\
        Sentiment-laden Words in Context & MFT & 1350 & Sentences that contain positive or negative words about movie-related sentiments (e.g. "I despise that director"). \\ 
        Temporal Sentiment Change & MFT & 2152 & Sentences that contain a change in sentiment over time regarding movies. Expectations are that depending on the final sentiment, the predictions are positive or negative. & used to, but now \\
        Negation of Positive Sentences & MFT & 1350 & Sentences with a negation of a positive sentiment. Predictions should be negative. & Simple negations: negative \\
        Negation of Positive, neutral words in the middle & MFT & 500 & Negation of a positive sentence, with a neutral sentiment in the middle. Should be negative. & Simple negations: not negative \\ 
        Movie Genre Specific Sentiments & MFT & 736 & Sentences with sentiments that are specific for a movie genre, such as horror movies being scary (positive). Based on the sentiment, prediction should be negative or positive. \\ 
        Movie Sentiments & MFT & 58 & Simple sentences about movies. Based on the sentiment, could be predicted as negative or positive.  \\
        Movie Industries Sentiments & MFT & 1200 & Sentences that talk about movie industries in a positive and negative way and the predictions should be accordingly. \\
        Change Neutral Words & INV & 500 / 3846 & Neutral words are changed with BERT, such as "that", "this", "of", etc. Expect the prediction to stay the same. & change neutral words with BERT \\
        Change Names & INV & 147 / 1617 & Changes the name in a review. Expectation is that the prediction stays the same. \\
        Negative names - Positive instances & INV & 157 / 1727 & Change names in positive reviews with names from training set that were only in negative reviews. Expectation is that predictions stay the same. & Polarizing Negative Names - Positive Instances \\
        Positive names - Negative instances & INV & 123 / 1353 & Change names in negative reviews with names from training set that were only in positive reviews. Expectation is that predictions stay the same. & Polarizing Positive names - Negative instances \\
        Negative names - Negative instances & INV & 123 / 1353 & Change names in negative reviews with names from training set that were only in negative reviews. Expectation is that predictions stay the same. & Polarizing Negative names - Negative instances \\
        Positive names - Positive instances & INV & 157 / 1727 & Change names in positive reviews with names from training set that were only in positive reviews. Expectation is that predictions stay the same. & Polarizing Positive names - Positive instances \\
        Change Movie Industries & INV & 18 / 252 & Changes a movie industry (Hollywood) to another one in movie reviews. After changing, the prediction should stay the same. \\
        Add Positive Phrases & DIR & 500 / 5500 & Add a positive phrase to a review, expecting the prediction confidence for positive to go up. \\
        Add Negative Phrases & DIR & 500 / 5000 & Add a negative phrase to a review, expecting the prediction confidence for negative to go up. \\
        \bottomrule
    \end{tabular}}
    \centering
    \caption{Overview of all the CheckList capabilities, with the test type, amount of examples, and a description of the capability provided. For clarity, we also provide the name of the capabilities originally used in the code, if they are not the same.}
    \label{tab:checklist_tests}
\end{table*}

\end{document}